\documentclass[10pt,twocolumn,letterpaper]{article}

\usepackage{wacv}
\usepackage{times}
\usepackage{epsfig}
\usepackage{graphicx}
\usepackage{amsmath}
\usepackage{amssymb}
\usepackage[accsupp]{axessibility}  

\usepackage{caption}
\usepackage{subcaption}
\usepackage{multirow}
\usepackage{tabularx}
\usepackage[ruled,vlined]{algorithm2e}
\usepackage{booktabs}

\def\wacvPaperID{****} 

\wacvfinalcopy 

\ifwacvfinal
\fi

\ifwacvfinal
\usepackage[breaklinks=true,bookmarks=false]{hyperref}
\else
\usepackage[pagebackref=true,breaklinks=true,colorlinks,bookmarks=false]{hyperref}
\fi

\begin{document}

\title{SSCAP: Self-supervised Co-occurrence Action Parsing \\ for Unsupervised Temporal Action Segmentation}

\author{Zhe Wang$^{1}$ \thanks{This work was done during internship at Amazon} \hspace{0.1cm} Hao Chen$^{2}$ \hspace{0.1cm} Xinyu Li$^{2}$ \hspace{0.1cm} Chunhui Liu$^{2}$ \hspace{0.1cm} \\ Yuanjun Xiong$^{2}$ \hspace{0.1cm} Joseph Tighe$^{2}$ \hspace{0.1cm} Charless Fowlkes$^{1,2}$ 
\\
\\
$^1$Dept. of CS, UC Irvine \hspace{2cm} $^2$Amazon Web Services 
}

\maketitle

\ifwacvfinal
\thispagestyle{empty}
\fi

\begin{abstract}
    Temporal action segmentation is a task to classify each frame in the video with an action label. However, it is quite expensive to annotate every frame in a large corpus of videos to construct a comprehensive supervised training dataset. Thus in this work we propose an unsupervised method, namely SSCAP, that operates on a corpus of unlabeled videos and predicts a likely set of temporal segments across the videos. SSCAP leverages Self-Supervised learning to extract distinguishable features and then applies a novel Co-occurrence Action Parsing algorithm 
  to not only capture the correlation among sub-actions underlying the structure of activities, but also estimate the temporal path of the sub-actions in an accurate and general way. We evaluate on both classic datasets (Breakfast, 50Salads) and the emerging fine-grained action dataset (FineGym) with more complex activity structures and similar sub-actions. Results show that SSCAP achieves state-of-the-art performance on all datasets
  and can even outperform some weakly-supervised approaches, demonstrating its effectiveness and generalizability.
  
\end{abstract}

\section{Introduction}

Temporal action segmentation aims to classify each frame in an untrimmed video with an action label. 
The task is a key step in understanding the structure of complex activities in videos. A recent study~\cite{shao2020tapos} also shows that the fine-grained sub-action segmentation can help many of other tasks such as action recognition and detection. Today's high performing solutions for temporal action segmentation require full frame-level supervision~\cite{mstcn,gcnactionseg,bcn}.  However, collecting per-frame annotations in untrimmed videos is extremely expensive and often impractical. 
Some recent work aim to overcome this by leveraging weak supervision to avoid the need of the full frame-level annotations~\cite{actionsetscvpr2020,setViterbi,nnviterbi}. These solutions assume that at least the number or the order of the sub-actions are given for each video. Even though they do not use full supervision, they still need some annotations that are costly to acquire, especially when expert knowledge on fine-grained atoms of activities (i.e., sub-actions) is required. To further reduce the annotation cost, we ask the question, can we do temporal action segmentation in a fully unsupervised way?

Unsupervised temporal action segmentation is a very challenging problem. 
To parse long video sequences into sub-actions in a semantically meaningful way without any supervision, it requires a solution to not only extract highly distinguishable visual representations for each individual frame, but also to capture the temporal relations among frames and sub-actions, so that the solution can well estimate the number and the order of the occurrence of each sub-action (i.e., the temporal path), which is typically provided in the weakly-supervised setting. The problem is even more challenging when dealing with videos that contain activities with complex structures and recurrence of sub-actions. Today there are still no effective solutions to handle these challenges 
and the performance of existing unsupervised temporal action segmentation solutions \cite{cvpr2019cnnrnn,kukleva2019unsupervisedtemporalembedding,cvpr2018} is still far from desired.

In this work, we propose a {\bf S}elf-{\bf S}upervised {\bf C}o-occurrence {\bf A}ction {\bf P}arsing method, namely SSCAP, for unsupervised temporal action segmentation. Specifically, we introduce self-supervised methods to extract features that are more temporal distinguishable. 
On top of it, we propose a Co-occurrence Action Parsing (CAP) algorithm to 
decode the frames into sub-actions by leveraging the estimated prior of the co-occurrence relations of sub-actions. The prior, though estimated over the whole dataset in an unsupervised manner, can statistically depict the natural structure of activities and hence help better estimate the number and the order of the occurrence of each sub-action 
during inference. Moreover, CAP takes recurrence of sub-actions into account in estimating the temporal path, and hence is able to handle complex structures of activities.  

Results show that SSCAP achieves state-of-the-art for unsupervised action segmentation, and even outperforms some of the weakly-supervised approaches. Besides evaluating on traditional action segmentation benchmarks such as Breakfast~\cite{breakfast}, Salad~\cite{salad}, we additionally test SSCAP on the more challenging recent dataset FineGym~\cite{shao2020FineGym}. 
Actions in traditional benchmarks are well structured - they are goal-directed (actors have clear goals like ``making salad'' or ``preparing scrambled egg'') and can be decomposed into clear steps, with one step being as the prerequisite for another one (e.g., you have to ``get'' an egg before ``scramble'' it). However, actions in FineGym have more complex structures - they are fast moving gymnastics actions, where several actors may perform the same sub-action multiple times in a single video. There are more similar fine-grained  sub-actions across different activities as well (for example: both `floor exercise' and `vault' events involve  a number of basic sub-action like `running' and `jumping'). Results show that comparing to the existing unsupervised action segmentation method, SSCAP can achieve significant performance improvements on this challenging dataset, demonstrating 
that SSCAP is especially superior in segmenting videos with complex activities and is more generalizable to different scenarios. To summarize, our main contributions are:

(1) We design a new solution called SSCAP to achieve temporal action segmentation in a fully unsupervised manner that can significantly reduce the annotation cost. SSCAP not only achieves state-of-the-art performance under the unsupervised setting, but also outperforms some weakly-supervised solutions;
    
(2) We introduce self-supervised learning in solving unsupervised action segmentation. While self-supervised learning has been widely exploited on other tasks such as image classification and action recognition~\cite{Deeo_Clustering,speednet,shufflelearn}, the application of it on action segmentation has been largely unexplored. We explore a few of today's most popular video based self-supervised learning approaches and show superior performance over the current state-of-the-art features in action segmentation. We further show that those self-supervised approaches that learn temporal characteristics outperform other self-supervised methods that focus on spatial transformation, which proves that modeling temporal relations is a key for accurately segmenting the frames into sub-actions;

(3) We propose a co-occurrence action parsing (CAP) algorithm, which takes into account the co-occurrence and recurrence of sub-actions calculated based on global statistics of all the videos, to estimate the temporal path of sub-actions in each video in a more precise and general way, and therefore outperforms today's solutions. Furthermore, 
we are the first to study action segmentation on FineGym, the emerging and more challenging dataset that contains more complex activities structures. 
Results show that the design of CAP achieves superior performance on this dataset compared to existing solutions.

\section{Related Work}
\label{section:relatedwork}

\textbf{Temporal action segmentation} is a task to classify each frame in an untrimmed video with a label of action. It is a key for understanding structures of complex activities. While current deep learning techniques~\cite{lin2019tsm,after-unet,Ji_2021_ICCV,feichtenhofer2016convolutional,Liu_2020_CVPR,li2021learning,cdg,ma2021transfusion,tran2015learning,gpa} have achieved great success on other action understanding tasks such as action recognition~\cite{i3d,feichtenhofer2019slowfast,yang2020temporal,wang2016temporal,zhou2018temporal,wang2018non,gmmhmm,zwicassp} and action detection~\cite{lin2018bsn,zhao2017temporal,xu2017r,hou2017tube}, temporal action segmentation is still far from being well-explored. Today's high performing solutions for temporal action segmentation require full frame-level supervision~\cite{mstcn,gcnactionseg,htk,bcn,lea2016segmental,lea2017temporal,ddxd}.
However, acquiring annotations on fine-grained sub-actions for each frame is expensive. To reduce the amount of labeled data needed,~\cite{sstda} proposes a domain adaptation approach based on self-supervised auxiliary tasks and achieves good performance. More recent works focus on developing weakly-supervised solutions: instead of having supervision for every frame, only the list (with order information)~\cite{firstunsuper,energyweaklyactionseg,nnviterbi,Richard_2017_CVPR,ding2018weakly,huang2016connectionist,chang2019d3tw} or the set (without order information)~\cite{actionsetscvpr2020,setViterbi,actionsetscvpr2018} of the sub-actions in each video are required during training. Other work uses script
or instructions for weak annotation~\cite{alayrac2016unsupervised}.
Unsupervised temporal action segmentation, which does not have any supervision signal per video, starts to receive growing interests these days, to further reduce the dependence on labeled data ~\cite{UnsuperProcLearn,cvpr2018,cvpr2019cnnrnn,kukleva2019unsupervisedtemporalembedding,vidalmata2020joint,JS,MNVALR}.  However, these solutions do not fully explore the self-supervised feature, do not take into account semantic correlations among sub-actions and the recurrence of sub-actions in some complex activity structures. The performance of them is still far from desired. 


\textbf{Self-supervised learning of visual representation} is a hot topic today, and has shown success in some of action understanding tasks~\cite{lstmunsupervisedfaeture,Deeo_Clustering,speednet,fernando2017self}.  Many different video properties have been used as supervision to build good representations, such as: cycle consistency between video frames \cite{xiaolongcvpr2019}; distinguishing between a video frame sequence and a shuffled version of it \cite{shufflelearn}; predicting the order of the shuffled clips \cite{cliporder}; solving space-time cubic puzzles \cite{spacetimecubicpuzzles}; classifying the speed of the video \cite{speednet,speednetMIT}; recognize the rotation of the video \cite{rotationet}. Another common task is predicting the future, either by predicting pixels of future frames \cite{motionnet}, or an embedding of a future video segment \cite{pdc}. While today's action segmentation solutions~\cite{mstcn,kukleva2019unsupervisedtemporalembedding,bcn} still mainly rely on hand-crafted features such as IDT~\cite{idt}, Fisher Vector (FV)~\cite{fishervector}, or pre-trained features such as I3D on Kinetics dataset~\cite{i3d}, we believe that introducing self-supervised methods in feature representation learning for unsupervised action segmentation can significantly boost the performance, as it can not only bring strong discriminative power to the features, but also bring a chance to improve the temporal characteristics of the features.

\begin{figure*}[htbp!]
\centerline{\includegraphics[scale=.45]{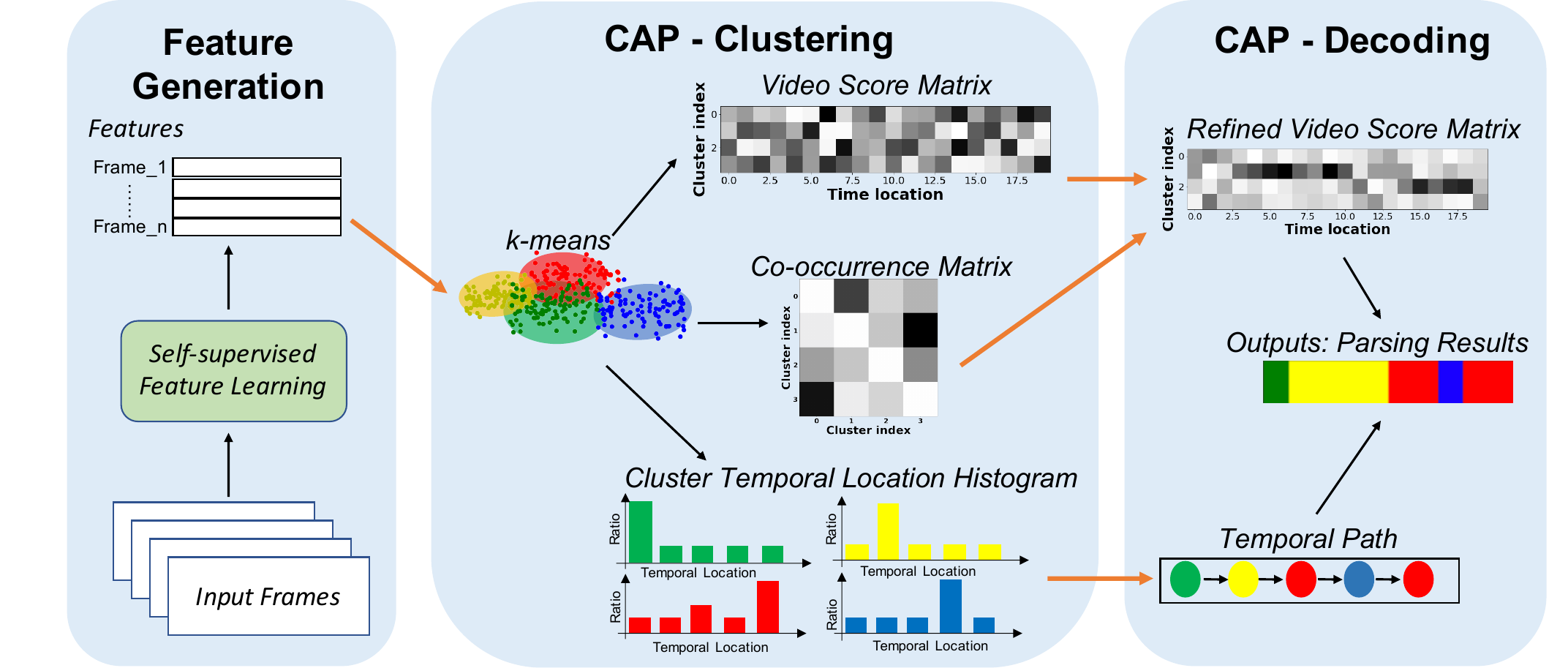}}
\caption{Overview of our SSCAP approach.}
\label{fig:flowchart}
\end{figure*}

\section{Technical Approach}
\label{section:method}
Temporal action segmentation is a task to classify each frame in the video to a sub-action. Specifically, given $M$ videos,  each of them is represented by $X_m$ which contains $N_m$ frames. The task is to predict the action label $l_{mn} \in {1,...,K}$ for each video frame $x_{mn}$. In this paper, we study the task in an unsupervised setting, i.e., the label $l_{mn}$ for each frame is unknown during model training.

Fig.~\ref{fig:flowchart} is an overview of our SSCAP approach. Briefly: given a set of videos, our feature generation module extracts frame features with a self-supervised representation learning method. Extracted frame features are then sent to the co-occurrence action parsing (CAP) module, which clusters the frames and decodes them into segments of unlabeled sub-actions. Finally, the semantic action label is assigned to each cluster by using Hungarian algorithm~\cite{hungarianalgorithm}. We introduce each module in details in the following sections.

\subsection{Feature Generation with Self-supervision}
\label{section:feature}

Our co-occurrence action parsing (CAP) module relies on the feature representation to have some semantic meaning so that the clustering step in CAP can effectively separate different semantic concepts in the video. In this work we explore a number of self-supervised feature leaning methods that have been developed for image and video classification tasks. We apply these methods for the task of video segmentation and study their relative effectiveness for that task v.s. the classification task they were designed for. Since one essential factor of the feature design for the temporal action segmentation is to capture the temporal characteristics, we want to further understand whether self-supervised learning methods built based on temporal augmentations (e.g, adjusting the frame rate, shuffling the frame or clip order, etc.) are also effective on temporal segmentation or not. 

With these questions in mind, we explore three of today's most popular video self-supervised learning methods and compare them with the feature designs used today in unsupervised action segmentation, including: IDT~\cite{idt},  FV~\cite{fishervector}, pre-trained I3D features~\cite{i3d}, and the continuous temporal features that achieves SOTA performance today~\cite{kukleva2019unsupervisedtemporalembedding}. Specifically, we study the following self-supervised methods: (1) \textit{SpeedNet}~\cite{speednet,speednetMIT}, which augments the frame rate of the video and predict it; (2) \textit{ShuffleLearn}~\cite{shufflelearn}, which predict whether the order of frames in the sequence is shuffled or not; (3) \textit{RotationNet} ~\cite{rotationet}, which rotates the whole video sequence by several pre-defined degrees and predict which degree the video is rotated. Note that RotationNet is a method based on spatial augmentation, while ShuffleLearn and SpeedNet are with temporal augmentation that are expected to have stronger temporal reasoning capability. Briefly we summarize the designs of these three methods in our work:

\textit{SpeedNet}~\cite{speednet,speednetMIT}. Given a clip $C$ with fixed number of $N$ frames: $C$ = ($f_1, f_2, ..., f_N$), we pre-define four frame rate settings $[2, 4, 8, 15]$. Each time when we generate a clip we randomly select one out of these four settings. Note that we keep the number of frames in each clip as the same, hence the time-range of the clip varies according to the frame rate selected. We train the network to predict which frame-rate the clip is sampled from (i.e., a 4-way classification).

\textit{ShuffleLearn}~\cite{shufflelearn}. Given a clip $C$ with fixed number of $N$ frames: $C$ = ($f_1, f_2, ..., f_N$), we shuffle $M$ frames ($0<M<N$) in a random order. Each time when we generate the clip, we flip the coin to decide whether to shuffle it or not. We train the network to predict whether a clip is shuffled or not.

\textit{RotationNet}~\cite{rotationet}. Given a clip $C$ with fixed number of $N$ frames: $C$ = ($f_1, f_2, ..., f_N$), we pre-define four rotation degree settings: $[0, 90, 180, 270]$. We also add some randomness when we rotate, i.e., in $[-30,30]$ degree. Each time when we generate clip we randomly select one out of these four settings and rotate the whole video sequence with the selected degree. We train the network to predict which of the settings is the rotation on (i.e., a 4-way classification).

\subsection{Co-occurrence Action Parsing}

Our co-occurrence action parsing (CAP) algorithm consists of two steps: clustering and decoding. Extracted frame-level features from Section~\ref{section:feature} are first clustered. Since we don't have ground truth annotations on sub-actions, here we aim at generating a number of clusters, with each of them representing a unique but unlabeled sub-action. After clustering, we then generate three outputs: (1) a {\it video score matrix} that represents the predicting score of each frame to each cluster for the whole video; 
(2) a {\it co-occurrence matrix} that captures the chance of any pair of clusters co-existing in videos; and (3) a {\it cluster temporal location histogram} that captures the distribution of temporal locations of each cluster in videos. In decoding, we first apply the co-occurrence matrix onto the video score matrix to generate a {\it refined video score matrix}. Then we estimate a {\it temporal path} with the cluster temporal location histogram and decode the frames to final segments based on the temporal path. The overall algorithm can be seen in Fig.~\ref{fig:flowchart}. Below we introduce the clustering and decoding in details.

\subsubsection{Clustering}
\label{sec:cluster}

Given an activity, and assuming there are $K$ different sub-actions ($K$ is known following the literature~\cite{kukleva2019unsupervisedtemporalembedding,cvpr2018}), 
we extract all the frame-level features for all videos labeled with this activity, and cluster them into $K$ clusters using k-means. To get the probability of the frame $x_{mn}$ that belongs to the cluster $k$, we estimate a Gaussian distribution for each cluster $k \in K$ :
\begin{equation}
\label{eq:prob}
p(x_{mn}|k) = \mathcal{N} (x_{mn} ;\mu_k,\Sigma_k)
\end{equation} 
We generate three outputs from the clustering step: (1) video score matrix; (2) co-occurrence matrix and (3) cluster temporal location histogram.  

\textbf{Video score matrix}. Given a video $X_m$ that contains $N_m$ frames, we create a video score matrix $S_m \in \mathbb{R}^{K \times N_m}$ to hold the probabilities calculated from Eq.~(\ref{eq:prob}), where $K$ is the number of clusters. Each element $S_m[k,n]$, with $k\in \{1,2,..K\}$ and $n \in \{1,2,...N_m\}$ represents the probability of the $n$-th frame assigned to the cluster $k$.

\textbf{Co-occurrence matrix}. The purpose of creating the co-occurrence matrix is to capture the correlations among sub-actions underlying the native structure of activities from the global video sets. Given an activity, there are usually strong correlations among its sub-actions across all videos. For example, in the activity ``Fried Egg'', ``fry egg'' always occurs with ``take egg''and ``crack egg''. While previous solutions do not build models to capture such global patterns, we leverage a co-occurrence matrix  to capture the co-existing relations among sub-actions. Note that, since we are dealing with an unsupervised setting, we don't have ground truth annotations on sub-actions, hence in the real implementation, what we calculate is the co-occurrence between two clusters $i$ and $j \in \{1,2, .., K\}$, instead of the real sub-actions. 
The way to generate the co-occurrence matrix is simple: for each pair of cluster $i$ and $j$, we iterate on all the videos and count the number of times that cluster $i$ and $j$ appear together in each video as $C(i,j)$. We also count the number of videos that contains the cluster $i$ as $C(i)$. 
We then calculate a conditional occurrence probability as $ P(j|i) = C(i,j)/C(i)$, which represents the probability of cluster $j$ appears in a video given that cluster $i$ appears.
We calculate $P(i|j)$ for all pairs of clusters $(i, j)$ in this way and generate the $K\times K$ dimension co-occurrence matrix. Details are shown in Algorithm~\ref{alg:selection}. Since the co-occurrence matrix is learned from all videos, it is capturing the global patterns of the activity structures. The co-occurrence matrix is used for refining the video score matrix, which is introduced in Section~\ref{sec:decode}.

\textbf{Cluster temporal location histogram}. The purpose of generating the cluster temporal location histogram is to estimate where each cluster generally locates in temporal dimension. Given the frames associated to each cluster from k-means, we compute the histogram of the temporal location for each cluster based on the timestamp of each frame in the video. Specifically, for a frame $x_{mn}$, i.e., the $n$-th frame in the video $X_m$, we calculate its relevant timestamp as $t_{x_{mn}} = n/N_m$, where $N_m$ is the total number of frames in video $X_m$. Then for each cluster $k$, we calculate the histogram on $t_{x_{mn}}$ for all the frames $x_{mn}$ from the videos that are clustered to $k$. The histogram indicates the potential temporal location of the cluster $k$ in the videos, and will be used for temporal path decoding introduced in Section~\ref{sec:decode}. Note that in real cases a sub-action can appear multiple times in different parts of videos, so each cluster might have more than one significant bin in the cluster temporal location histogram.

\subsubsection{Decoding}
\label{sec:decode}

After clustering all the frames into $K$ clusters, the next step is to decode them into segments of the sub-actions. There are two main steps, namely, (1) refine the video score matrix with the co-occurrence matrix; (2) estimate the temporal path of the clusters and decode frames to segments.

\textbf{Refining the video score matrix.} The first step in the decoding is to generate the refined video score matrix $R_m \in \mathbb{R}^{K \times N_m}$ for each video $X_m$. There are two main issues with directly using the original video score matrix $S_m$ in decoding: (1) the video score matrix is generated purely based on the frame feature representations, it does not capture any global patterns of the activity structures across all videos, and does not contain any correlation information among sub-actions; (2) the clustering is conducted on $K$ clusters, which is the total number of different sub-actions in the given activity. In many cases, only a subset of $K$ will appear in each individual video due to the diversity of the activities being performed by different persons. 
Directly using video score matrix may lead to the issue of over-segmenting video to the non-existing classes in current video. 

To address these issues, we apply the co-occurrence matrix $P$ to refine the score matrix $S_m$. Specifically,  
given a video $X_m$ and for each cluster $k$, we first calculate the number of frames belonging to $k$ in the video, i.e., $c(k)$, and define the ratio of frames for cluster $k$ as $r(k)= c(k)/N_m$, where $N_m$ is the total number of frames in the video.

Then, in each iteration, we pick the cluster $k^*$ with the current largest ratio of frame $r(k)$ and update the video scores $S_m[j,:]$ for the remaining clusters $j$ (i.e., $r(j) \leq r(k^*)$) with the corresponding co-occurrence probability $P(j|k^*)$. Then we pop out $k^*$ and repeat the process for the remaining clusters. Details are in Algorithm~\ref{alg:selection}. By doing so we add global information from the whole dataset gradually into each individual video. Scores of clusters that are not strongly correlated with the dominating clusters in this video will be attenuated. The probabilistic update is inspired by Kalman filter where the new probability $R_m[j,:]$ is updated based on the observation $P(j|k^*)$ and old probability $S_m[j,:]$. 

\begin{algorithm}[t]
\small
  \SetAlgoLined
  \KwData{Video score matrix $S_m$ for video $X_m$, with $S_m[k,n]_{k=1,n=1}^{K,N_m}$, number of clusters: $K$.} 
  \KwResult{Co-occurrence matrix: $P(i,j)_{i,j=1}^K$,
  Refined video score matrix $R_m$, with scores $R_m[k,n]$.}
  \textbf{\textit{Generate Co-occurrence Matrix}}\\ 
  - Iterate all the videos, count the times each cluster appears $C(i)_{i=1}^K$, and the times different clusters co-occur together $C(i,j)_{i,j=1}^K$. Normalize to make it as conditional
  probability $P(j|i) = C(i,j) / C(i)$\\
  \textbf{\textit{Refined Video Score Matrix}}\\ 
  - Initialization:  $\mathcal{G} \leftarrow k_0 $ ($k_0$ is the cluster with the largest ratio of frames  $r(k_0)$ in current video).\\
  - $k^* = k_0$ \\
  \While{$len(\mathcal{G}) \leq K$ and  $r(k^*) \textgreater 0$}{
    1. For each remaining cluster $j\not\in \mathcal{G}$: \\
        - update the video score matrix conditioned on the previous selected cluster $k^*$:
    $
       R_m[j,n] = P(j | k^*) \cdot S_m[j,n]
    $\\
    2. Select the next cluster: $k^* \leftarrow \mathrm{arg}\max_{j} r(j)  $. \\
    3. Update: $\mathcal{G} \leftarrow \mathcal{G} \cup \{k^*\}$
    }
  - Return: $P, R_m$.
  \caption{Generating the Co-occurrence Matrix and the Refined Video Score Matrix.}
  \label{alg:selection}
\end{algorithm}

\textbf{Estimating the temporal path and decoding.} Having the refined video score matrix $R_m$, the final step is to decode the frame sequences into the sub-action segments. Assume that we have a known temporal path of the clusters $k_1 \rightarrow ... \rightarrow k_T$, then the decoding is to maximize the probability of the sequence following this path based on our refined video score matrix, i.e.,  
\begin{equation}
\label{eq:decode}
\hat{l}_{l}^{N_m} = {argmax}_{l_1,...,l_{N_m}}\prod^{N_m}_{n=1} p(x_{mn}|l_n ) p(l_n | l_{n-1}),
\end{equation} 
where, $p(x_{mn}|l_n=k) = R_m$ is the probability of the frame $x_{mn}$ belonging to the cluster $k$, obtained from the refined video score matrix, and $p(l_n | l_{n-1}) $ are the transition probabilities of moving from the label  $l_{n-1}$  at frame $n-1$ to the next label $l_n$ at frame $n$. Given a temporal path, Eq.~(\ref{eq:decode}) can be solved with a Viterbi algorithm \cite{viterbi}. 

The temporal path estimation is the key for the decoding. While in weakly-supervised action segmentation the temporal path (i.e, the order of sub-actions) is usually assumed to be known, in unsupervised setting it can only be estimated, which is challenging. 
Moreover, many of literature work today assume a strict single directional temporal path with each sub-action can appear at most once in the path. Such an assumption may suit for dataset with simple activity structures, e.g., Breakfast~\cite{breakfast}, however, begins to fail when the structure of activities gets more complex. For instance, in some activities, a single sub-action may occur multiple times, and the order of the sub-actions can be bi-directional (i.e., sub-action $i$ can go to sub-action $j$ and then go back to $i$). Such sub-action recurrence brings further difficulties to precisely estimate the temporal path. 

We design a {\it multi-occur temporal path} estimation algorithm that is applicable to both simple and complex cases,
using the {\it cluster temporal location histogram} introduced in Section~\ref{sec:cluster}. 
For each cluster, we select the bins from temporal location histogram with counts (normalized to the total counts across all the bins) larger than a certain value.
Then we concatenate the selected bins from all the clusters and order them into a time sequence based on the temporal location of each bin, which is calculated as $T_i = \sum{t_{x_{mn}^i}} / N_i$, where $t_{x_{mn}^i}$ is the relvant timestamp of frame $x_{mn}^i$ that fall into bin $i$, and $N_i$ is the total number of frames in bin $i$. The order of bins with their corresponding cluster labels forms our temporal path. Our multi-occur temporal path 
is able to capture the multi-time occurrence for each cluster as well as the bi-direction transition among different clusters, because more than one bins from each cluster can be selected. 

\section{Experiments}
\label{section:experiments}

\paragraph{Datasets.} We evaluate our approach on two most widely used datasets in action segmentation: Breakfast~\cite{breakfast} and 50Salads~\cite{salad}, and an emerging fine-grained action dataset FineGym~\cite{shao2020FineGym}. \textbf{Breakfast}~\cite{breakfast} is a large-scale dataset that consists of ten different complex activities 
of performing common kitchen activities with approximately eight sub-actions
per activity class. The duration of the videos varies significantly, e.g. `coffee' has an average duration of 30 seconds while `cooking pancake' takes roughly 5 minutes. 
\textbf{50Salads}~\cite{salad} dataset contains 4.5 hours of different people performing a single complex activity, making mixed
salad. Compared to the other datasets, the videos are much longer with an average video length of 10k frames. We also observe that a single sub-action can occur multiple times along the video in this dataset, which seldom happens in Breakfast. \textbf{FineGym}~\cite{shao2020FineGym} is an emerging dataset provided with fine-grained sports action annotations. Though it is originally proposed for fine-grained action recognition, it contains rich hierarchical information including annotations of start and end for the fine-grained sub-actions, which is suitable for action segmentation evaluation.  These sports videos contain several athletes performing the same action 
and their replays. The structure of the videos are more complex and it includes many recurrences of a single sub-action, and includes more similar fine-grained sub-actions (up to 35), which makes the dataset more challenging for action segmentation than typical ones used today. We use Gym99 setting in original paper.

 \begin{table}[t]
 {
 \centering
 \begin{tabular}{@{} c|c|c @{}} 
  \hline
   \textit{Breakfast} & MoF & F1 score\\ 
   \hline
   \multicolumn{3}{l}{\textbf{Unsupervised setting}}      \\
   \hline
         GMM \cite{cvpr2018}&  0.346 & - \\
         LSTM + AL \cite{cvpr2019cnnrnn}&  0.429* & - \\
         CTE \cite{kukleva2019unsupervisedtemporalembedding} &  0.418 & 0.264   \\ 
        VTE-UNET \cite{vidalmata2020joint} & 0.481 & - \\ 
        ASAL \cite{JS} & 0.525 & 0.379 \\
       \textbf{Our SSCAP} &  0.511  &  \textbf{0.392} \\ 
       \hline
       \multicolumn{3}{l}{\textbf{Weakly-supervised setting}}      \\
   \hline
         Action Sets \cite{actionsetscvpr2018} & 0.284  & - \\
         NNviterbi \cite{nnviterbi}&  0.430  & - \\
         SCT \cite{actionsetscvpr2020} & 0.304  & - \\
         SetViterbi \cite{setViterbi} &  0.408 & -   \\ 
          EnergySeg \cite{energyweaklyactionseg} &  0.630  & -   \\ 
       \hline
       \multicolumn{3}{l}{\textbf{Fully-supervised setting}}      \\
   \hline
         HTK \cite{htk} & 0.259  & - \\
         GTRM \cite{gcnactionseg} & 0.650& - \\
         MS-TCN \cite{mstcn}&  0.663 & - \\
         BCN \cite{bcn} &  0.704  & -   \\ 
       \hline
 \end{tabular}
 \caption{Results on Breakfast.} 
 \label{tab:actionsegmentationBreakfast}
 }
 \end{table}

\paragraph{Evaluation Metrics.} 
We use the Hungarian algorithm \cite{hungarianalgorithm} to get a one-to-one matching that maximizes the similarity between all the clusters and sub-action classes for all the videos in evaluation. The number of total sub-actions within each dataset is known and used to define the number of clusters K.
We report frame-wise accuracy as the \textit{mean over frames (MoF)}, indicating the percentage of frames correctly labelled.  
We also report \textit{F1 score} the same as \cite{kukleva2019unsupervisedtemporalembedding} to measure the quality of the temporal segmentation.

\subsection{Comparison with SOTA}

We compare our SSCAP with SOTA solutions on Breakfast in Table~\ref{tab:actionsegmentationBreakfast}, 50Salads in Table~\ref{tab:actionsegmentationSalads} and FineGym in Table~\ref{tab:actionsegmentationFineGym}. For all the datasets, We use SpeedNet~\cite{speednet} trained on the Kinetics-400~\cite{i3d} as the self-supervised feature in our SSCAP. Details on feature comparison can be found in Section~\ref{sec:abfeature}. We can see SSCAP achieves SOTA performance on both Breakfast and 50Salads datasets in the unsupervised setting.
SSCAP on Breakfast even outperforms most of the weakly-supervised solutions.

 \begin{table}[t]
 {
 \centering
 \begin{tabular}{@{} c|c|c @{}} 
  \hline
   \textit{50Salads} & MoF &F1 score\\ 
   \hline
   \multicolumn{3}{l}{\textbf{Unsupervised setting}}      \\
   \hline
         LSTM + AL \cite{cvpr2019cnnrnn}&  0.606* &  - \\
         CTE \cite{kukleva2019unsupervisedtemporalembedding} &  0.355 &  -   \\
         VTE-UNET \cite{vidalmata2020joint} & 0.306 & - \\
         ASAL \cite{JS} & 0.392 & - \\
       \textbf{Our SSCAP} &  \textbf{0.414} &  \textbf{0.303} \\ 
       \hline
       \multicolumn{3}{l}{\textbf{Weakly-supervised setting}}      \\
   \hline
         NNviterbi \cite{nnviterbi}&  0.494 &  - \\
          EnergySeg \cite{energyweaklyactionseg} &  0.547  & -   \\ 
       \hline
       \multicolumn{3}{l}{\textbf{Fully-supervised setting}}      \\
   \hline
         HTK \cite{htk} & 0.247 & \\
         GTRM \cite{gcnactionseg} & 0.826 & - \\
         MS-TCN \cite{mstcn}&  0.734 &  - \\
         BCN \cite{bcn} &  0.844 & -  \\ 
       \hline
 \end{tabular}
 \caption{Results on 50Salads. 
 Note that `*' uses a per-video instance level ``cluster to ground-truth'' mapping, which uses additional video-level information, while we follow other work to use ``cluster to ground-truth'' mapping in a global manner in evaluation and do not use any additional information~\cite{vidalmata2020joint}.
 }

 \label{tab:actionsegmentationSalads}
 }
 \end{table}

 \begin{table}[t]
 {
 \centering
 \begin{tabular}{ @{} c|c|c @{}} 
  \hline 
   \textit{FineGym} & MoF & F1 score   \\ 
 \hline  
       Baseline~\cite{kukleva2019unsupervisedtemporalembedding} &  0.294  &  0.167   \\  
       \textbf{Our SSCAP}  &  \textbf{0.666} & \textbf{0.297}   \\  
       \hline
 \end{tabular}
 \caption{Results on FineGym.}
 \label{tab:actionsegmentationFineGym}
 }
 \end{table}

Since FineGym is new and we are the first to conduct action segmentation on it, there is no existing related results yet. We generate the baseline by running the SOTA apporach~\cite{kukleva2019unsupervisedtemporalembedding} on FineGym. We can see that on this challenging dataset, SSCAP achieves significant improvement 
on both MoF and F1 compared to the baseline solution, demonstrating the effectiveness of our designed solution in handling videos with more complex structures. Overall the consistent improvement across all the benchmarks shows that our approach is able to generalize to different cases. 

\subsection{Ablation Study on SSCAP} 

Next, we conduct ablation studies on SSCAP. Three main units are investigated: (1) Self-supervised Feature Learning; (2) Co-occurrence Matrix; (3) Multi-occur Temporal Path. Results are summarized in Table~\ref{tab:betterdecoding}. 

\textbf{Effectiveness of Self-supervised Learning.} From Table 4 we observe that applying self-supervised feature learning consistently improves performance over the baseline using continuous temporal features~\cite{kukleva2019unsupervisedtemporalembedding}, on all of the three datasets, demonstrating the value of self-supervised feature learning on the action segmentation task.

\textbf{Effectiveness of Co-occurrence Matrix.}
We evaluate the influence of using the co-occurrence matrix to refine the video score matrix in Table~\ref{tab:betterdecoding} on all datasets. From the table, our co-occurrence matrix brings performance improvement over using original video score on all of the three datasets and on both \textit{MoF} and \textit{F1}. 
The improvement on FineGym dataset is more notable than the other two, indicating the importance of using the co-occurrence matrix while handling more complex scenarios. 

\begin{table}[t]
{
\begin{center}
\begin{tabular}{@{}  c | c  c c  | c | c @{}} 
 \hline 
 Dataset   & SS & C-Matrix & M-T-Path & MoF & F1  \\ 
\hline  
 \multirow{4}{*}{\textit{Breakfast}}  &     &  &    &  0.418  & 0.264  \\  
    &  \checkmark  &   &    &  0.508  & 0.391  \\  
      
      & \checkmark &    \checkmark  &   
 &  0.511  & 0.392  \\  
        
         &  \checkmark &  \checkmark  &   \checkmark  &  \textbf{0.511}  &  \textbf{0.392} \\  
       \hline  
    \multirow{4}{*}{\textit{50Salads}} &      &   &    &  0.355  & -  \\  
    &   \checkmark   &   &    &  0.372  & 0.281  \\  
      
        & \checkmark &  \checkmark  &    &  0.378  &  0.290 \\  
        
          &   \checkmark  & \checkmark  & \checkmark  &  \textbf{0.414}  & \textbf{0.303}  \\

        \hline  
    \multirow{4}{*}{\textit{FineGym}}      &          & &  &  0.294  &  0.167 \\  
    &      \checkmark    & &  &  0.425  &  0.246 \\  
      
         &  \checkmark &  \checkmark  &   &  0.442  &  0.248 \\  
        
          &   \checkmark   & \checkmark &  \checkmark  & 
          \textbf{0.666}  &  \textbf{0.297} \\  

      \hline  
\end{tabular}
\end{center}
\caption{Ablation study on all datasets. `SS' denotes whether to use the \textit{self-supervised feature learning} (SpeedNet feature trained on K400); `C-Matrix' denotes whether to use the  \textit{co-occurrence matrix} to refine the video score matrix; and `M-T-Path' denotes whether to use our designed \textit{multi-occur temporal path} in decoding. If none of them is selected then it's a baseline setting with the continuous temporal feature~\cite{kukleva2019unsupervisedtemporalembedding}.}
\label{tab:betterdecoding}
}
\end{table}

\textbf{Effectiveness of Multi-occur Temporal Path.} We then assess the impact of using the proposed multi-occur temporal path in decoding. Results on all three datasets are in Table~\ref{tab:betterdecoding}.  
We can see the designed multi-occur temporal path decoding improves both \textit{MoF} and \textit{F1} on 50Salads and FineGym notably, but does not on Breakfast.
We found that the recurrence of sub-actions in a video happens more frequently in 50Salads and FineGym, while in Breakfast, most of the sub-actions only occur once in each video, therefore, the multi-occur temporal path is not as helpful on Breakfast as on the other datasets. The improvement on FineGym is significant, representing that multi-occur temporal path is superior in segmenting activities with complex structures. It is worth to mention that on the Breakfast dataset, our multi-occur temporal path algorithm generates exactly the same temporal path (with each sub-action occurring only once) automatically as the path generated from turning off the multi-occur algorithm, representing that our algorithm is self-applicable to different scenarios.

\subsection{Comparison of Feature Learning} 
\label{sec:abfeature}

In the section, we explore what type of self-supervised design best fits for the unsupervised action segmentation task. We study three different self-supervised feature learning methods: (1) SpeedNet~\cite{speednet}, (2) RotationNet~\cite{rotationet}, and (3) ShuffleLearn~\cite{shufflelearn}, and compare the results on Breakfast dataset in Table~\ref{tab:differentmodelssinglecolumn}. All the numbers in the table are generated without using the co-occurrence matrix and the multi-occur temporal path, so that we can purely evaluate the effectiveness of features without bringing any additional factors.   

We conduct the experiments on self-supervised training (1) only on Kinetics dataset (i.e., row (d) - (f)), (2) only on Breakfast dataset (i.e., row (g) - (i)) and (3) first on Kinectics dataset and then on Breakfast dataset (i.e., row (j) - (l)), to further explore that whether we need to train on target dataset (i.e., Breakfast in our case) to get a good feature representation for the task. From the table we can find that: 

(1) Based on experiment (a) - (l), we conclude that SpeedNet trained on Kinectics shown in row (d) performs the best; 

(2) Based on experiment (d) - (l), RotationNet consistently performs worse than SpeedNet and ShuffleLearn. This may due to the fact that RotationNet is using self-supervised signal from spatial dimension, which may lack of ability to model the temporal characteristics between frames. SpeedNet and ShuffleLearn, however, use temporal augmentation as the self-supervised signal, so that they may have stronger capability in modeling the temporal relations among frames and thus are more suitable for temporal action segmentation. The results also indicate that not all the self-supervised learning methods can help improve the performance, it is important to capture temporal relations among frames in feature learning for the segmentation task. In addition, SpeedNet, as one of the most emerging video self-supervised learning approaches, consistently outperforms the ShuffleLearn, showing the effectiveness of it; 

(3) Based on experiment (b), (d), (e), (f), self-supervised features always perform better than classical I3D feature pre-trained on Kinetics. We argue that 
while the I3D feature trained for action recognition task optimizes feature representation on video clip level, self-supervised learning representation implicitly learns better frame-level representation and can better capture the temporal structure inside the clip;

(4) Based on the experiments (d) - (l), we can see larger dataset like Kinetics can help build better self-supervised representation, while smaller ones contain less variety. It's not needed to use target dataset to get a good feature representation for the temporal action segmentation task.

\begin{table}[t]
{
\centering
\begin{tabular}{@{} c|c|c|c@{}} 
 \hline
   & Feature Description  & MoF & F1\\ 
   \hline
   \multicolumn{4}{l}{\textbf{Baseline}}      \\
   \hline
  (a) &  IDT \cite{kukleva2019unsupervisedtemporalembedding} &      0.316 &    -   \\ 
  (b) &     K400 I3D  \cite{kukleva2019unsupervisedtemporalembedding}&      0.251 &  -     \\ 
  (c) &     CTE  \cite{kukleva2019unsupervisedtemporalembedding} &      0.418 &    0.264   \\ 
  \hline
  \multicolumn{4}{l}{\textbf{Self-supervised on K400}}      \\
   \hline
  (d) &    K400 SpeedNet &      \textbf{0.508} &    \textbf{0.391}   \\ 
  (e) &     K400 RotationNet &       0.328 &     0.317   \\ 
  (f) &      K400 shuffleLearn &    0.339    &   0.328     \\ 
  \hline
  \multicolumn{4}{l}{\textbf{Self-supervised on Breakfast}}      \\
   \hline
  (g) &     Breakfast SpeedNet &     0.344 &    0.327   \\ 
  (h) &      Breakfast  RotationNet&      0.307 &  0.319     \\ 
  (i) &      Breakfast shuffleLearn &      0.315 &      0.309 \\ 
  \hline
  \multicolumn{4}{l}{\textbf{Self-supervised first on K400, then on Breakfast}}      \\
   \hline
  (j) &      K400, Breakfast, SpeedNet &      \underline{0.501} &    \underline{0.337}   \\ 
  (k) &     K400, Breakfast, RotationNet &      0.279 &  0.290     \\ 
  (l) &      K400, Breakfast, shuffleLearn &      0.292 &    0.318   \\ 

 \hline
\end{tabular}
\caption{Comparison of features for temporal action segmentation on the Breakfast \cite{breakfast} dataset.  `K400' denotes features trained on Kinetics \cite{i3d} dataset while `Breakfast' represents features trained on Breakfast \cite{breakfast} dataset. `SpeedNet', `RotationNet' and `ShuffleLearn' denote features trained using the self-supervised methods from \cite{speednet}, \cite{rotationet} and \cite{shufflelearn}. }
\label{tab:differentmodelssinglecolumn}
}
\end{table}

\subsection{Qualitative Results}

\begin{figure}[t]
\centering
\begin{subfigure}[b]{0.48\textwidth}
   \centering
  \includegraphics[width=.95\linewidth]{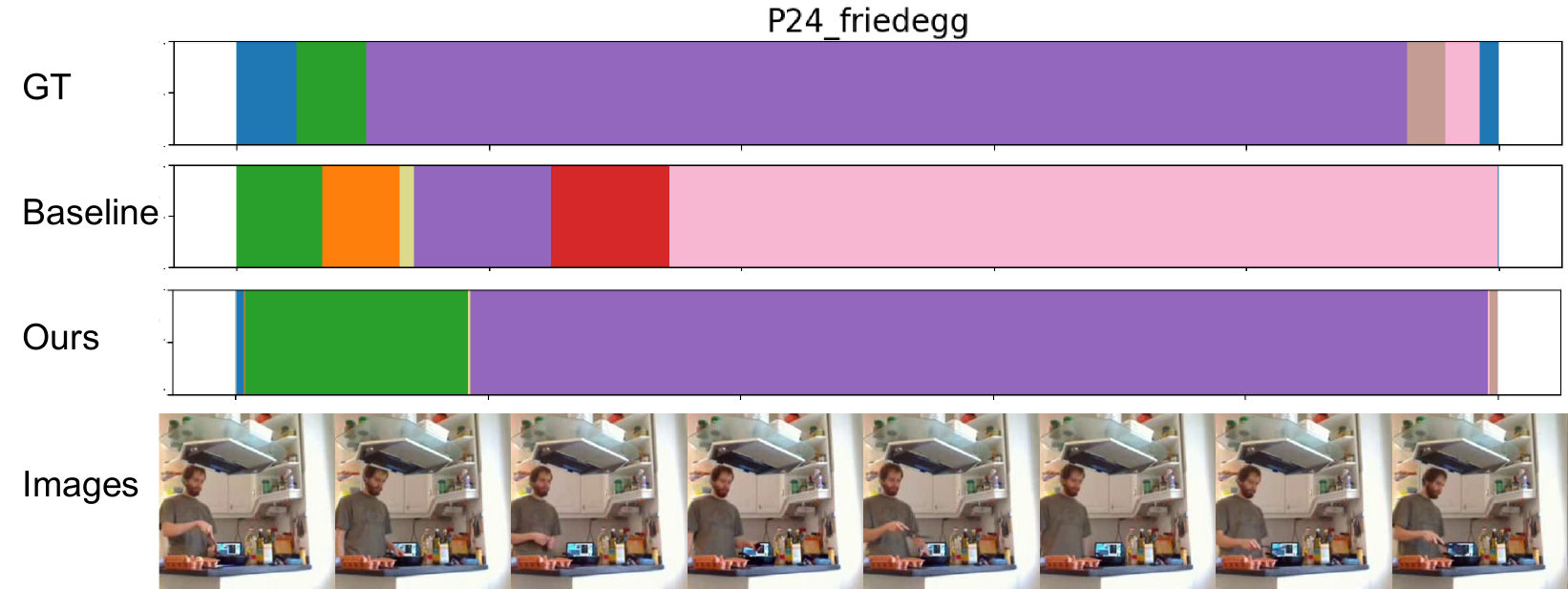} 
  \caption{P24 friedegg sequence from Breakfast dataset. The semantics order are `background', `crack egg',`fry egg', `take plate', `put egg2plate', `background'.}
  \label{fig:qualitativec}
\end{subfigure}
\hfill
\begin{subfigure}[b]{0.48\textwidth}
   \centering
  \includegraphics[width=.95\linewidth]{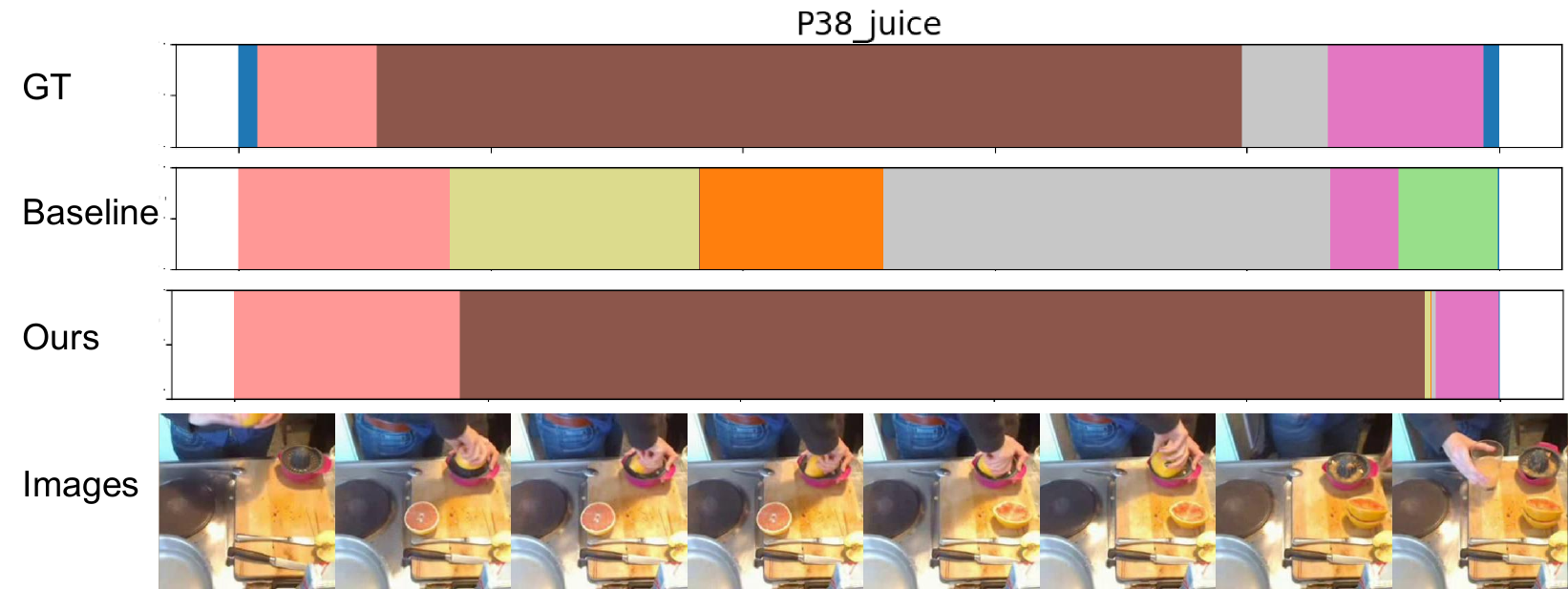}
  \caption{P38 juice sequence from Breakfast dataset. The semantics order are `background', `cut orange', `squeeze orange',`take glass',`pour juice',`background'.}
  \label{fig:qualitatived}
\end{subfigure}
\caption{Two qualitative examples (a) and (b). The first row is the frame-wise ground-truth (x-axis stands for the time in the video while different sub-action classes are colorized differently); the second row is the prediction from \cite{kukleva2019unsupervisedtemporalembedding}; the third row is our prediction; and the fourth row is the uniform sampled 8 frames from the original video. Best viewed zoom in and in color.}
\label{fig:qualitativeall}
\end{figure}

Fig.~\ref{fig:qualitativeall} visualizes two examples from the Breakfast dataset. From both Fig.~\ref{fig:qualitativec} and Fig.~\ref{fig:qualitatived}, we can see that SSCAP is able to  notably improve the segmentation quality. We can also see that the CAP algorithm is able to effectively suppressing the over-segmentation issue by introducing the co-occurrence relations among sub-actions in decoding.  

\section{Conclusion}
\label{section:concolusion}
In this work, we have proposed SSCAP, an unsupervised temporal action segmentation solution that uses self-supervised methods in feature learning 
and a co-occurrence action parsing algorithm that helps model the correlation among sub-actions and better handle complex activity structures in videos. 
SSCAP has achieved SOTA performance on three public benchmarks in unsupervised setting, and has even outperformed several recently proposed weakly-supervised methods. The future direction can be leveraging the deep and hierarchical clustering approaches in further improving the accuracy of the method.

\section*{Acknowledgement}
We'd like to give our special thanks to Fanyi Xiao,  Bing Shuai and Uta Büchler for their helpful discussions.

{\small
\bibliographystyle{ieee_fullname}
\bibliography{egbib}

\begin{thebibliography}{10}\itemsep=-1pt

\bibitem{cvpr2019cnnrnn}
Sathyanarayanan~N. Aakur and Sudeep Sarkar.
\newblock A perceptual prediction framework for self supervised event
  segmentation.
\newblock In {\em CVPR}, 2019.

\bibitem{alayrac2016unsupervised}
Jean-Baptiste Alayrac, Piotr Bojanowski, Nishant Agrawal, Josef Sivic, Ivan
  Laptev, and Simon Lacoste-Julien.
\newblock Unsupervised learning from narrated instruction videos.
\newblock In {\em CVPR}, 2016.

\bibitem{speednetMIT}
Sagie Benaim, Ariel Ephrat, Oran Lang, Inbar Mosseri, William~T. Freeman,
  Michael Rubinstein, Michal Irani, and Tali Dekel.
\newblock Speednet: Learning the speediness in videos.
\newblock In {\em CVPR}, 2020.

\bibitem{firstunsuper}
Piotr Bojanowski, Rémi Lajugie, Francis Bach, Ivan Laptev, Jean Ponce,
  Cordelia Schmid, and Josef Sivic.
\newblock Weakly supervised action labeling in videos under ordering
  constraints.
\newblock In {\em ECCV}, 2014.

\bibitem{lstmunsupervisedfaeture}
Biagio Brattoli, Uta Büchler, Anna-Sophia Wahl, Martin~E. Schwab, and Björn
  Ommer.
\newblock Lstm self-supervision for detailed behavior analysis.
\newblock In {\em CVPR}, 2017.

\bibitem{Deeo_Clustering}
Mathilde Caron, Piotr Bojanowski, Armand Joulin, and Matthijs Douze.
\newblock Deep clustering for unsupervised learning of visual features.
\newblock In {\em ECCV}, 2018.

\bibitem{i3d}
Joao Carreira and Andrew Zisserman.
\newblock Quo vadis, action recognition? a new model and the kinetics dataset.
\newblock In {\em CVPR}, 2017.

\bibitem{chang2019d3tw}
Chien-Yi Chang, De-An Huang, Yanan Sui, Li Fei-Fei, and Juan~Carlos Niebles.
\newblock D3tw: Discriminative differentiable dynamic time warping for weakly
  supervised action alignment and segmentation.
\newblock In {\em CVPR}, 2019.

\bibitem{sstda}
Min-Hung Chen, Baopu Li, Yingze Bao, Ghassan AlRegib, and Zsolt Kira.
\newblock Action segmentation with joint selfsupervised temporal domain
  adaptation.
\newblock In {\em CVPR}, 2020.

\bibitem{gmmhmm}
George~E. Dahl, Dong Yu, Li Deng, and Alex Acero.
\newblock Context-dependent pre-trained deep neural networks for
  large-vocabulary speech recognition.
\newblock In {\em IEEE Transactions on Audio, Speech, and Language Processing},
  2012.

\bibitem{motionnet}
Ali Diba, Vivek Sharma, Luc~Van Gool, and Rainer Stiefelhagen.
\newblock Dynamonet: Dynamic action and motion network.
\newblock In {\em Arxiv}, 2019.

\bibitem{ding2018weakly}
Li Ding and Chenliang Xu.
\newblock Weakly-supervised action segmentation with iterative soft boundary
  assignment.
\newblock In {\em CVPR}, 2018.

\bibitem{UnsuperProcLearn}
E. {Elhamifar} and Z. {Naing}.
\newblock Unsupervised procedure learning via joint dynamic summarization.
\newblock In {\em ICCV}, 2019.

\bibitem{speednet}
Dave Epstein, Boyuan Chen, and Carl Vondrick.
\newblock Oops! predicting unintentional action in video.
\newblock In {\em CVPR}, 2020.

\bibitem{mstcn}
Yazan~Abu Farha and Juergen Gall.
\newblock Ms-tcn: Multi-stage temporal convolutional network for action
  segmentation.
\newblock In {\em CVPR}, 2019.

\bibitem{actionsetscvpr2020}
Mohsen Fayyaz and Juergen Gall.
\newblock Sct: Set constrained temporal transformer for set supervised action
  segmentation.
\newblock In {\em CVPR}, 2020.

\bibitem{feichtenhofer2019slowfast}
Christoph Feichtenhofer, Haoqi Fan, Jitendra Malik, and Kaiming He.
\newblock Slowfast networks for video recognition.
\newblock In {\em ICCV}, 2019.

\bibitem{feichtenhofer2016convolutional}
Christoph Feichtenhofer, Axel Pinz, and Andrew Zisserman.
\newblock Convolutional two-stream network fusion for video action recognition.
\newblock In {\em CVPR}, 2016.

\bibitem{fernando2017self}
Basura Fernando, Hakan Bilen, Efstratios Gavves, and Stephen Gould.
\newblock Self-supervised video representation learning with odd-one-out
  networks.
\newblock In {\em CVPR}, 2017.

\bibitem{rotationet}
Spyros Gidaris, Praveer Singh, and Nikos Komodakis.
\newblock Unsupervised representation learning by predicting image rotations.
\newblock In {\em ICLR}, 2018.

\bibitem{pdc}
Tengda Han, Weidi Xie, and Andrew Zisserman.
\newblock Video representation learning by dense predictive coding.
\newblock In {\em ICCV}, 2019.

\bibitem{hou2017tube}
Rui Hou, Chen Chen, and Mubarak Shah.
\newblock Tube convolutional neural network (t-cnn) for action detection in
  videos.
\newblock In {\em ICCV}, 2017.

\bibitem{huang2016connectionist}
De-An Huang, Li Fei-Fei, and Juan~Carlos Niebles.
\newblock Connectionist temporal modeling for weakly supervised action
  labeling.
\newblock In {\em ECCV}, 2016.

\bibitem{gcnactionseg}
Yifei Huang, Yusuke Sugano, and Yoichi Sato.
\newblock Improving action segmentation via graph based temporal reasoning.
\newblock In {\em CVPR}, 2020.

\bibitem{Ji_2021_ICCV}
Pan Ji, Runze Li, Bir Bhanu, and Yi Xu.
\newblock Monoindoor: Towards good practice of self-supervised monocular depth
  estimation for indoor environments.
\newblock In {\em ICCV}, 2021.

\bibitem{spacetimecubicpuzzles}
Dahun Kim, Donghyeon Cho, and In~So Kweon.
\newblock Self-supervised video representation learning with space-time cubic
  puzzles.
\newblock In {\em AAAI}, 2019.

\bibitem{breakfast}
Hilde Kuehne, Ali Arslan, and Thomas Serre.
\newblock Recovering the syntax and semantics of goal-directed human
  activities.
\newblock In {\em CVPR}, 2014.

\bibitem{htk}
Hilde Kuehne, Juergen Gall, and Thomas Serre.
\newblock An end-to-end generative framework for video segmentation and
  recognition.
\newblock In {\em WACV}, 2016.

\bibitem{hungarianalgorithm}
H.~W. Kuhn.
\newblock The hungarian method for the assignment problem.
\newblock In {\em Wiley Online Library}, 1955.

\bibitem{kukleva2019unsupervisedtemporalembedding}
Anna Kukleva, Hilde Kuehne, Fadime Sener, and Jurgen Gall.
\newblock Unsupervised learning of action classes with continuous temporal
  embedding.
\newblock In {\em CVPR}, 2019.

\bibitem{lea2017temporal}
Colin Lea, Michael~D Flynn, Rene Vidal, Austin Reiter, and Gregory~D Hager.
\newblock Temporal convolutional networks for action segmentation and
  detection.
\newblock In {\em CVPR}, 2017.

\bibitem{lea2016segmental}
Colin Lea, Austin Reiter, Ren{\'e} Vidal, and Gregory~D Hager.
\newblock Segmental spatiotemporal cnns for fine-grained action segmentation.
\newblock In {\em ECCV}, 2016.

\bibitem{energyweaklyactionseg}
Jun Li and Sinisa Todorovic.
\newblock Weakly supervised energy-based learning for action segmentation.
\newblock In {\em ICCV}, 2019.

\bibitem{setViterbi}
Jun Li and Sinisa Todorovic.
\newblock Set-constrained viterbi for set-supervised action segmentation.
\newblock In {\em CVPR}, 2020.

\bibitem{JS}
Jun Li and Sinisa Todorovic.
\newblock Action shuffle alternating learning for unsupervised action
  segmentation.
\newblock In {\em CVPR}, 2021.

\bibitem{li2021learning}
Runze Li, Srikrishna Karanam, Ren Li, Terrence Chen, Bir Bhanu, and Ziyan Wu.
\newblock Learning local recurrent models for human mesh recovery, 2021.

\bibitem{lin2019tsm}
Ji Lin, Chuang Gan, and Song Han.
\newblock Tsm: Temporal shift module for efficient video understanding.
\newblock In {\em ICCV}, 2019.

\bibitem{lin2018bsn}
Tianwei Lin, Xu Zhao, Haisheng Su, Chongjing Wang, and Ming Yang.
\newblock Bsn: Boundary sensitive network for temporal action proposal
  generation.
\newblock In {\em ECCV}, 2018.

\bibitem{Liu_2020_CVPR}
Wenqian Liu, Runze Li, Meng Zheng, Srikrishna Karanam, Ziyan Wu, Bir Bhanu,
  Richard~J. Radke, and Octavia Camps.
\newblock Towards visually explaining variational autoencoders.
\newblock In {\em CVPR}, 2020.

\bibitem{ma2021transfusion}
Haoyu Ma, Liangjian Chen, Deying Kong, Zhe Wang, Xingwei Liu, Hao Tang, Xiangyi
  Yan, Yusheng Xie, Shih-Yao Lin, and Xiaohui Xie.
\newblock Transfusion: Cross-view fusion with transformer for 3d human pose
  estimation.
\newblock In {\em BMVC}, 2021.

\bibitem{shufflelearn}
Ishan Misra, C.~Lawrence Zitnick, and Martial Hebert.
\newblock Shuffle and learn: Unsupervised learning using temporal order
  verification.
\newblock In {\em ECCV}, 2016.

\bibitem{fishervector}
Florent Perronnin, Jorge Sánchez, and Thomas Mensink.
\newblock Improving the fisher kernel for large-scale image classification.
\newblock In {\em ECCV}, 2010.

\bibitem{viterbi}
T. Quach and M Farooq.
\newblock Maximum likelihood track formation with the viterbi algorithm.
\newblock In {\em IEEE Conference on Decision and Control}, 1994.

\bibitem{Richard_2017_CVPR}
Alexander Richard, Hilde Kuehne, and Juergen Gall.
\newblock Weakly supervised action learning with rnn based fine-to-coarse
  modeling.
\newblock In {\em CVPR}, 2017.

\bibitem{actionsetscvpr2018}
Alexander Richard, Hilde Kuehne, and Juergen Gall.
\newblock Action sets: Weakly supervised action segmentation without ordering
  constraints.
\newblock In {\em CVPR}, 2018.

\bibitem{nnviterbi}
Alexander Richard, Hilde Kuehne, Ahsan Iqbal, and Juergen Gall.
\newblock Neuralnetwork-viterbi: A framework for weakly supervised video
  learning.
\newblock In {\em CVPR}, 2018.

\bibitem{MNVALR}
M.~Saquib Sarfraz, Naila Murray, Vivek Sharma, Ali Diba, Luc~Van Gool, and
  Rainer Stiefelhagen.
\newblock Temporally-weighted hierarchical clustering for unsupervised action
  segmentation.
\newblock In {\em CVPR}, 2021.

\bibitem{cvpr2018}
Fadime Sener and Angela Yao.
\newblock Unsupervised learning and segmentation of complex activities from
  video.
\newblock In {\em CVPR}, 2018.

\bibitem{shao2020FineGym}
Dian Shao, Yue Zhao, Bo Dai, and Dahua Lin.
\newblock Finegym: A hierarchical video dataset for fine-grained action
  understanding.
\newblock In {\em CVPR}, 2020.

\bibitem{shao2020tapos}
Dian Shao, Yue Zhao, Bo Dai, and Dahua Lin.
\newblock Intra- and inter-action understanding via temporal action parsing.
\newblock In {\em CVPR}, 2020.

\bibitem{salad}
Sebastian Stein and Stephen~J. McKenna.
\newblock Combining embedded accelerometers with computer vision for
  recognizing food preparation activities.
\newblock In {\em UBICOMP}, 2013.

\bibitem{tran2015learning}
Du Tran, Lubomir Bourdev, Rob Fergus, Lorenzo Torresani, and Manohar Paluri.
\newblock Learning spatiotemporal features with 3d convolutional networks.
\newblock In {\em ICCV}, 2015.

\bibitem{vidalmata2020joint}
Rosaura~G. VidalMata, Walter~J. Scheirer, Anna Kukleva, David Cox, and Hilde
  Kuehne.
\newblock Joint visual-temporal embedding for unsupervised learning of actions
  in untrimmed sequences.
\newblock In {\em WACV}, 2021.

\bibitem{ddxd}
Dong Wang, Di Hu, Xingjian Li, and Dejing Dou.
\newblock Temporal relational modeling with self-supervision for action
  segmentation.
\newblock In {\em AAAI}, 2021.

\bibitem{idt}
Heng Wang and Cordelia Schmid.
\newblock Action recognition with improved trajectories.
\newblock In {\em ICCV}, 2013.

\bibitem{wang2016temporal}
Limin Wang, Yuanjun Xiong, Zhe Wang, Yu Qiao, Dahua Lin, Xiaoou Tang, and
  Luc~Van Gool.
\newblock Temporal segment networks: Towards good practices for deep action
  recognition.
\newblock In {\em ECCV}, 2016.

\bibitem{wang2018non}
Xiaolong Wang, Ross Girshick, Abhinav Gupta, and Kaiming He.
\newblock Non-local neural networks.
\newblock In {\em CVPR}, 2018.

\bibitem{xiaolongcvpr2019}
Xiaolong Wang, Allan Jabri, and Alexei~A Efros.
\newblock Learning correspondence from the cycle-consistency of time.
\newblock In {\em CVPR}, 2019.

\bibitem{gpa}
Zhe Wang, Liyan Chen, Shauray Rathore, Daeyun Shin, and Charless Fowlkes.
\newblock Geometric pose affordance: 3d human pose with scene constraints.
\newblock In {\em Arxiv}, 2019.

\bibitem{bcn}
Zhenzhi Wang, Ziteng Gao, Limin Wang, Zhifeng Li, and Gangshan Wu.
\newblock Boundary-aware cascade networks for temporal action segmentation.
\newblock In {\em ECCV}, 2020.

\bibitem{cdg}
Zhe Wang, Daeyun Shin, and Charless Fowlkes.
\newblock Predicting camera viewpoint improves cross-dataset generalization for
  3d human pose estimation.
\newblock In {\em ECCVW}, 2020.

\bibitem{zwicassp}
Zhe Wang, Yali Wang, Limin Wang, and Yu Qiao.
\newblock Codebook enhancement of vlad representation for visual recognition.
\newblock In {\em ICASSP}, 2016.

\bibitem{cliporder}
Dejing Xu, Jun Xiao, Zhou Zhao, Jian Shao, Di Xie, and Yueting Zhuang.
\newblock Self-supervised spatiotemporal learning via video clip order
  prediction.
\newblock In {\em CVPR}, 2019.

\bibitem{xu2017r}
Huijuan Xu, Abir Das, and Kate Saenko.
\newblock R-c3d: Region convolutional 3d network for temporal activity
  detection.
\newblock In {\em ICCV}, 2017.

\bibitem{after-unet}
Xiangyi Yan, Hao Tang, Shanlin Sun, Haoyu Ma, Deying Kong, and Xiaohui Xie.
\newblock After-unet: Axial fusion transformer unet for medical image
  segmentation.
\newblock In {\em WACV}, 2022.

\bibitem{yang2020temporal}
Ceyuan Yang, Yinghao Xu, Jianping Shi, Bo Dai, and Bolei Zhou.
\newblock Temporal pyramid network for action recognition.
\newblock In {\em CVPR}, 2020.

\bibitem{zhao2017temporal}
Yue Zhao, Yuanjun Xiong, Limin Wang, Zhirong Wu, Xiaoou Tang, and Dahua Lin.
\newblock Temporal action detection with structured segment networks.
\newblock In {\em ICCV}, 2017.

\bibitem{zhou2018temporal}
Bolei Zhou, Alex Andonian, Aude Oliva, and Antonio Torralba.
\newblock Temporal relational reasoning in videos.
\newblock In {\em ECCV}, 2018.

\end{thebibliography}
}

\end{document}